\documentclass[runningheads]{llncs}

\usepackage[utf8]{inputenc}
\usepackage[english]{babel}
\usepackage{graphicx}
\usepackage{enumitem}
\usepackage{amsmath, amssymb, mathrsfs}
\usepackage{url}
\usepackage{color}
\usepackage[style=american]{csquotes}
\usepackage{booktabs}
\usepackage{todonotes}

\begin{document}

\title{Understanding and Preparing Data of Industrial Processes for Machine Learning Applications}
\titlerunning{Preparing Data of Industrial Processes for Machine Learning Applications}

\footnotetext[1]{The final publication is available at \url{https://link.springer.com/chapter/10.1007\%2F978-3-030-45093-9\_50}} 

\author{Philipp Fleck\inst{1} \and
Manfred Kügel\inst{2} \and
Michael Kommenda\inst{1}}
\authorrunning{P. Fleck et al.}

\institute{
	Heuristic and Evolutionary Algorithms Laboratory \\
	University of Applied Sciences Upper Austria \\
	Softwarepark 11, 4232 Hagenberg i.M., Austria \and
	Primetals Technologies Austria GmbH \\
	Turmstraße 44, 4031 Linz, Austria
	\\
    \medskip
	\email{\{philipp.fleck, michael.kommenda\}@fh-hagenberg.at}
	\email{manfred.kuegel@primetals.com}
}

\maketitle

\begin{abstract}
	Industrial applications of machine learning face unique challenges due to the nature of raw industry data. 
	Preprocessing and preparing raw industrial data for machine learning applications is a demanding task that often takes more time and work than the actual modeling process itself and poses additional challenges.
	This paper addresses one of those challenges, specifically, the challenge of missing values due to sensor unavailability at different production units of nonlinear production lines.
	In cases where only a small proportion of the data is missing, those missing values can often be imputed.
	In cases of large proportions of missing data, imputing is often not feasible, and removing observations containing missing values is often the only option.
	This paper presents a technique, that allows to utilize all of the available data without the need of removing large amounts of observations where data is only partially available.
	We do not only discuss the principal idea of the presented method, but also show different possible implementations that can be applied depending on the data at hand.
	Finally, we demonstrate the application of the presented method with data from a steel production plant.

\end{abstract}

\section{Introduction}

\begin{figure}[t]
	\centering
	\includegraphics[page=1, width=0.7\textwidth, clip, trim=0cm 12cm 18cm 0cm]{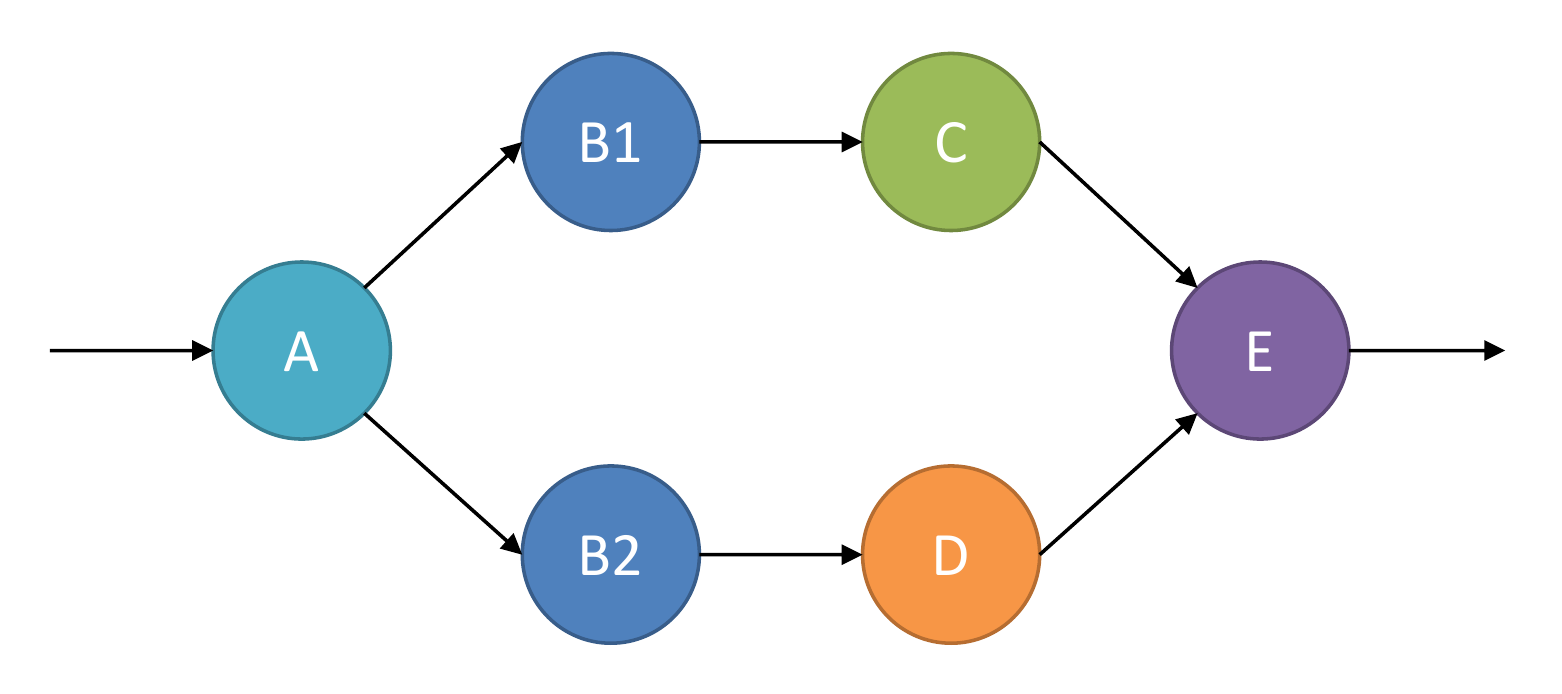}
	\label{fig:production_line_sample}
	\caption{A nonlinear production layout where an item passes through different processing units. As a result, sensor data for a specific item is only available if it passes a certain processing unit.}
\end{figure}

In industrial production processes, items are often processed in nonlinear production layouts with different sensor types equipped at different production stages, leading to sparse sensor data caused by the different routes items take through the production facility. 
Considering the sample production layout in Fig. \ref{fig:production_line_sample}, all processed items will pass station A and E; therefore, data from those station will always be available. 
Station B1 and B2 are of the same type with the same sensors, thus, data from B is always available, regardless if an item was processed at B1 or B2. However, depending on the route, only sensor data from station C or D is available, never both.

Most machine learning algorithms struggle with sparse data due to their inability to handle missing values themselves.
For such cases, missing values are often \emph{imputed} \cite{royston2004multiple} with the help of existing data.
Imputing missing values is often feasible in scenarios, where missing data occurs infrequently and at random.
However, imputing missing values can be challenging if large proportions of the data is missing, especially if related signals that could be used to reconstruct the missing values are also missing at the same time.
This is often the case, because related signals from the same production unit are also completely missing if the product did not pass this production unit.
Instead, observations or even entire signals that contain missing values are often removed due to the lack of alternatives \cite{garcia2015data}.
However, considering a production line with two exclusive paths, removing observations that contain any missing values would lead to the removal of the entire data.
Alternatively, removing all signals for the exclusive paths, leads to a considerable loss of information.
As a conclusion, neither imputing nor removing missing values are desirable options.

\begin{figure}[ht]
	\centering
	\includegraphics[page=2, width=0.7\textwidth, clip, trim=0cm 9.5cm 19cm 0cm] {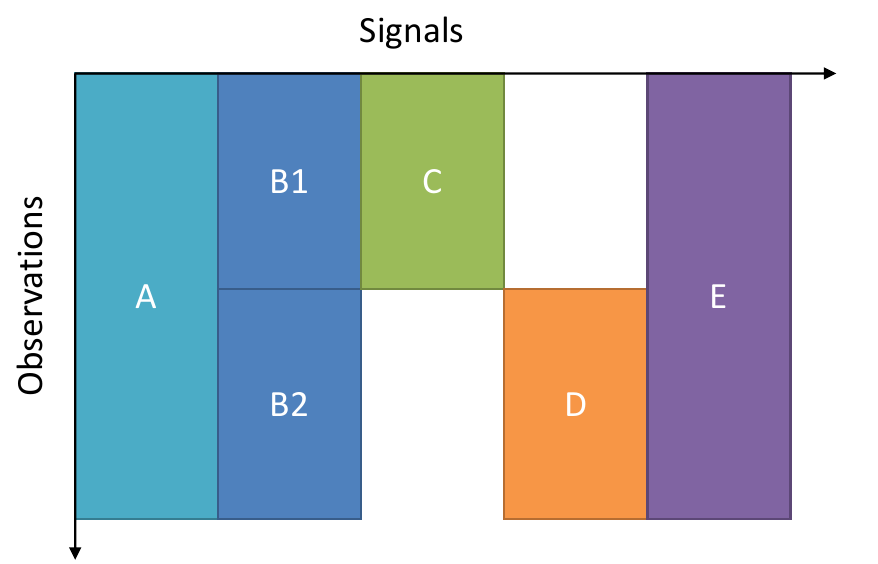}
	\caption{This chart shows available data, grouped by the processing units based on the production layout in Fig. \ref{fig:production_line_sample}. }
	\label{fig:missing_values_overview}
\end{figure}

Fig. \ref{fig:missing_values_overview} shows an overview of available and missing data for the sample production layout from Fig. \ref{fig:production_line_sample}.
The full block A symbolizes that this data is always available for all observations, which is also true for E. 
B is always available, either in the form of B1 or B2. 
C is only available for the top half of the observations and D only for the lower half. 
Note that the grouping of C and D in the top and bottom half was done for illustration, this is not necessarily true in real world cases.
If we remove either column C or D, potentially valuable signals are lost. 
Alternatively, if we remove all observations where D is not available, then half the available data from A, B, and E is lost.

Some machine learning algorithms are capable of dealing with missing values on their own \cite{tierney2015using}.
For example, random forests can be extended, so that for each decision within the a tree, there is a default choice in case a missing value is at hand.
In this paper, however, we present a method on a higher level, that can be used in conjunction with different modeling techniques that are not required to handle missing values themselves.

As an exemplary use case for this paper, we use the data of a production plant producing steel coils, where missing values in the data are caused by the different routes that the coil takes through the production plant.
For this use case, we will demonstrate and discuss in detail how we applied the proposed methods.

\section{Methods}

We propose a method for dealing with missing values due to nonlinear production lines, that can utilize all available data for training a model, without the need of imputation or removal of missing values.
This method consists of the following three steps:
\begin{enumerate}[label={(\arabic*)}]
	\item Create multiple subsets of the data that do not contain any missing values.
	\item Create an ensemble model \cite{zhou2012ensemble}, based on the different data subsets.
	\item Predict, using only models of the ensemble, where the corresponding input variables are available.
\end{enumerate}

The basic idea of the proposed method is to create multiple, missing-value-free datasets based on the actual dataset.
Then, multiple models are trained on these new datasets.
When a prediction is made, only models are selected where all corresponding input variables are available, and the model results are combined.
Typically, we create a base model based on features that are always available, and multiple residual models that are used as additional correction factors.
In other words, the residual models are building upon the predictions of the base model, and do only predict the error of the base model.
This way, the residual models can reuse knowledge the base model has already discovered.

\begin{figure}
	\centering
	\includegraphics[page=3, width=1.0\textwidth, clip, trim=0cm 5.5cm 10cm 0cm]{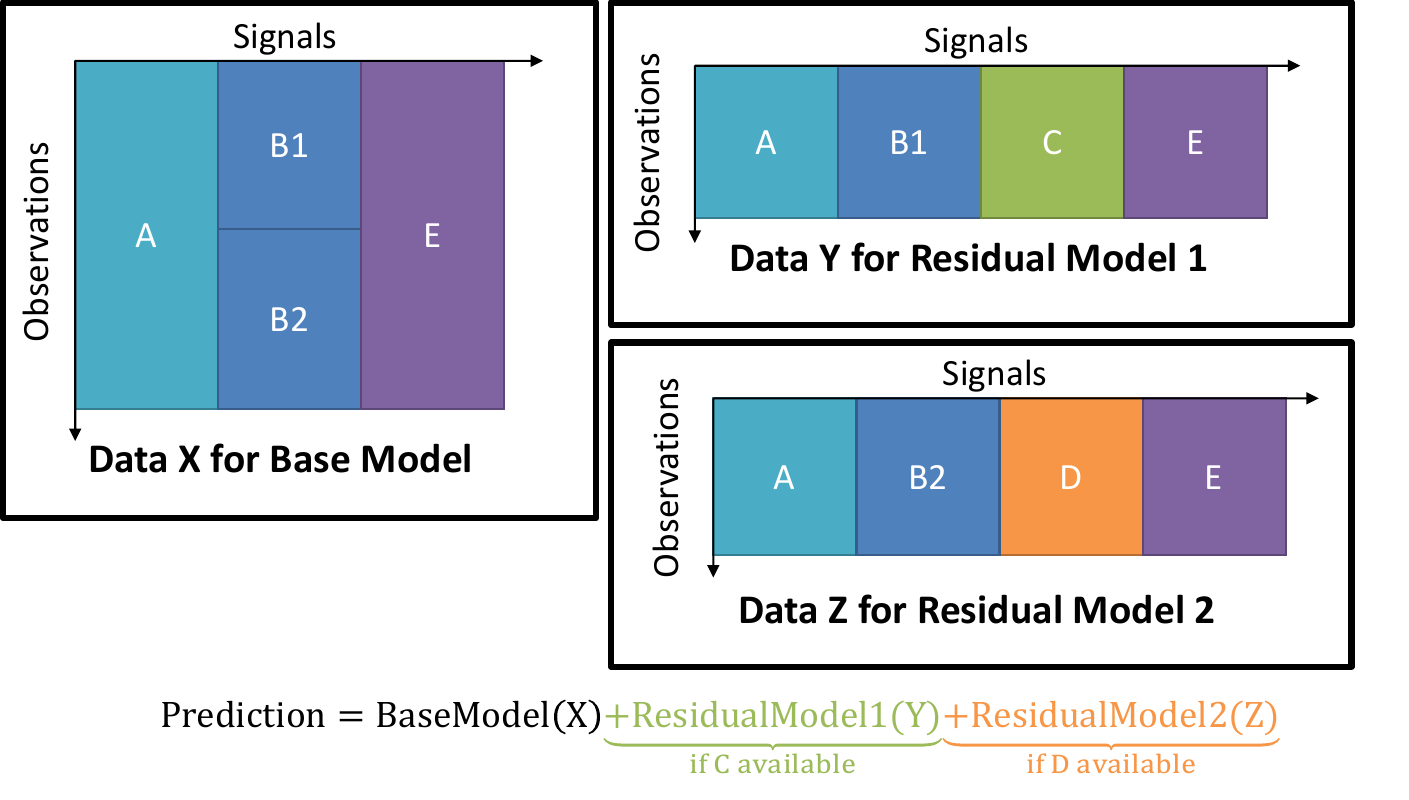}
	\caption{An overview and example of our proposed method that splits the data into multiple subparts, trains an ensemble model on those subparts, and uses selected models of the ensemble for a final prediction.}
	\label{fig:method_overview}
\end{figure}

An overview and example of this method is shown in Fig. \ref{fig:method_overview}.
Typically, we create a base model based on signals that are always available, for instance, signals A, E and, B (which is the combination of B1 and B2).
The data for residual model 1 only contains observations where C is available, as well as A, B1 and E for those observations. 
Likewise, the data for residual model 2 contains only observations where D is available.
To obtain a prediction value of a new observation, the base model is always evaluated while the application of the residual models 1 and 2 is depending on the availability of C and D.

The challenge of this method is to anticipate, which signals will commonly be available together, so a data subset can be created accordingly.
Additionally, all three steps are interdependent and also dependent on the data, therefore selecting appropriate strategies is important.
In the following sections, we describe each section in detail and also discuss advantages and disadvantages of the presented strategies for each step.

\subsection{Data Subset Strategies}

The first step is to create subsets of the dataset that do not contain any missing values.
Those subsets define which models of the ensemble are available, therefore, should be selected based on signals that are commonly available together.
As a base, there are two common strategies that we encountered:

\begin{description}
	\item[Subset by Grouped Signals:] 
	In this strategy, the data is split based on signals from the same processing unit.
	This is done, for example, in Fig. \ref{fig:method_overview}, where we split whether C or D is available, thus we obtain datasets for a base model and for two residual models.
	For this strategy there are two options concerning which signals are included in the data subset:
	If relations between the signals for a single processing unit (C or D) and the other data (A, B, or C) are suspected, both should be included.
	If no relationship is to be expected, only the signals from the processing unit (C or D) should be used in the data subset.
	\item[Subset by Common Routes:]
	If the production layout graph is complex and contains many branches, creating all potential combinations of data subsets would result in a large number of subsets and residual models.
	Instead, it is advised to split the data by common production routes.
	For example, even if the production layout would allow for a high number of potential routes in theory, there is often only a small number of routes that are used predominantly.
	In such case, the data can be split along those routes with the signals used that are available for the specific routes.
	Uncommon routes can either be grouped together with only their common signals remaining, or completely dropped.
	We will demonstrate such a splitting strategy later in Section \ref{sec:Application}.
\end{description}

\subsection{Ensemble Modeling}

For modeling, there are two considerations: the actual modeling method and the aggregation of the individual models' results.
As for modeling methods, there are no restrictions; thus, different methods can be used, also for the separate models.
For aggregating the models' results, there are two common strategies for ensembles:
\begin{description}
	\item[Bagging] (bootstrap aggregating) \cite{breiman1996bagging} is combining all models equally, e.g. averaging all predictions for a regression ensemble.
	In this case, each model predicts the target itself, and all applicable models' outputs are simply averaged.
	A downside of this approach is that models cannot built upon results that are already achieved by other models.
	\item[Boosting] involves an incremental modeling method that builds upon the results of previous models \cite{friedman2002stochastic}.
	Instead of each model trying to predict the target itself, a model only tries to predict the residuals from a previous model, allowing the residual models to build upon knowledge that the base model (or previous residual models) has already discovered.
	This is a core element of our proposed method, because it allows the residual models to build upon knowledge that was created from different data than the residual model has available. 
	This additional data corresponds to the observations that were removed from the original data when building the residuals data subset due to missing values and, therefore, would not be available if the model was trained in a conventional manner.
\end{description}


\subsection{Ensemble Prediction}

When a prediction is made based on the ensemble model, only models of the ensemble are used where all the input variables are available.
In other words, based on the available data, all applicable models are selected, and their prediction combined.
In case of a bagging model, the models' average is determined; in case of a boosting model, the base model is evaluated and the residual models' predictions (i.e. the correction factors) are added.

\section{Application}
\label{sec:Application}

\begin{figure}[h]
	\includegraphics[page=5, width=1.0\textwidth, clip, trim=0.2cm 2.8cm 10.3cm 0cm]{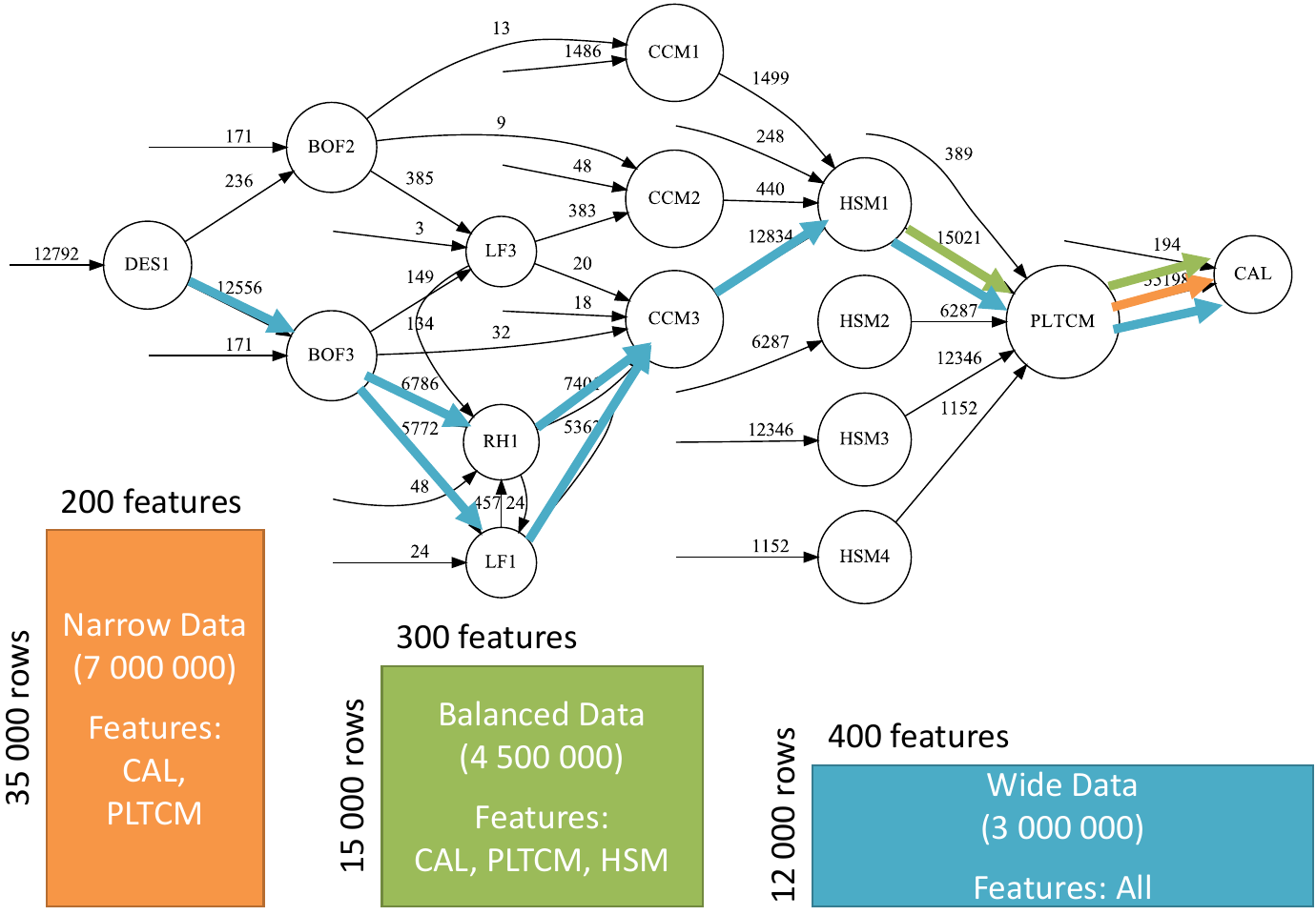}
	\caption{Layout with route-segments for splitting the data. The numbers in the graph show the number of coils that passed this route. On the bottom, the subsets of the data are shown with their number of values and included signal groups.}
	\label{fig:Routes}
\end{figure}

In this section, we demonstrate how we applied the proposed method, especially how we partitioned the data into missing-value-free subsets, and which ensemble modeling strategy we applied.
As an exemplary use case for this paper, we use the data of a production plant producing steel coils, where missing values are caused by different routes through the production plant, shown in Fig. \ref{fig:Routes}.

For building data subsets, one option would be to split the data by processing unit, e.g. one partition for all processing units of the same type (one for BOF, one for CCM, etc.).
However, since there is a single predominated route that most of the coils followed, we opted for splitting the data by common routes.
More specifically, we split by route-segments that a steel coil must pass in order to be considered for a data subset.
For example, we define a short route-segment (orange in Fig \ref{fig:Routes}) only spanning PLTCM and CAL, and all all coils that passed this route-segment are selected for our first data subset, which we call \emph{narrow data}.
As for the signals, only PLTCM and CAL signals are used.
Next, the green route, forming the route-segment HSM1-PLTCM-CAL is used for the \emph{balanced data}, which contains less coils than the previous route (i.e. coils from HSM2-4 are removed), but additionally also contain HSM signals.
Only HSM1 is used for this route-segment, because there is no sensor data available for the other HSMs.
Finally, the blue route, spanning the entire production layout, is used for the \emph{wide data} and contains even less coils than the other routes, but in exchange, all signals are available.
Although there is a branch within the blue route that would usually lead to missing values, we consider RH and LF to be identical since they both have the same signals.

For the ensemble modeling, we opted for a boosting strategy with incremental residual models.
As a start, we use the \emph{narrow data} that contains the most observation as a base model.
Next, the first residual model uses the \emph{balanced data} and predicts the residuals from the base model.
And finally, a second residual model uses the \emph{wide data} and predicts the residuals of the first residual model.
This sequential application of the residual model is possible, because all the signal of a model's dataset are also available in the subsequent model's dataset, i.e. all signals of the \emph{narrow data} are also available at the \emph{balanced data}.

As a result, if a coil only passed PLTCM and CAL, the base model is used for prediction.
In case a coil also passed HSM1, a correcting term based on the first residual model is applied.
In case the coil passed the long, blue route from DES1 to CAL, also the second residual model is applied.

Table \ref{tab:Results} shows the accuracy of the ensemble model of our proposed method, and a conventional model as comparison where observations containing missing values were removed.
We can clearly observe that our method yields higher accuracy (in terms of lower mean absolute error and higher $R^2$) than the conventional model.
Since the individual modeling methods had the same inputs and observations available, the results indicate, that the proposed method was successful in creating better models due to its ability to incorporate data that would otherwise has been removed.

\begin{table}
	\centering
	\caption{Model accuracies on the test partition for our proposed method and a conventional model were observations with missing values were removed.}
\begin{tabular}{l|ll|ll|ll}
\toprule
 & \multicolumn{2}{|c}{Narrow} & \multicolumn{2}{|c}{Balanced} & \multicolumn{2}{|c}{Wide} \\ 
 & \multicolumn{1}{|c}{MAE} & \multicolumn{1}{c}{$R^2$} &\multicolumn{1}{|c}{MAE} & \multicolumn{1}{c}{$R^2$} & \multicolumn{1}{|c}{MAE} &\multicolumn{1}{c}{$R^2$} \\
\midrule
Proposed Method 	& 0.285 & 0.549 & 0.233 & 0.717 & 0.183 & 0.811 \\
Conventional Model	& 0.285 & 0.549 & 0.284 & 0.585 & 0.315 & 0.450 \\
\bottomrule
\end{tabular} 
	\label{tab:Results}
\end{table}

For this paper, the main focus was the application of the proposed method.
Thus, only minimal parameter optimization and method tuning was performed.
However, both the proposed method and the conventional method used the same training data as well as same algorithm parameters for a fair comparison.

\section{Conclusion}

The presented method enables working with data that contains many missing values if they do occur systematically.
One typical case is industry data, where the production layout is nonlinear and therefore, a single item often does not pass all sensors.

The idea of our method is to build subsets of the data without missing values, based on signals that are available together.
Then, we train a model for each data subset, similar to ensemble modeling.
For a prediction with the ensemble model, the applicable models are selected based on the available data and the models' required input features, and then the predictions of the models are combined.
We showed an application using a boosting approach, where we build residual models based on common routes within the production layout.

Although the methodology was designed for industrial applications of machine learning, the presented method is also applicable in general if missing values occur.
However, the method strongly relies on an underlying reason why groups of signals are missing together, for example due to sensors only available at certain processing units.
In cases where missing values occur infrequently and at random, the proposed method is not recommended. 

In this paper, we have proposed a new method for dealing with missing values that appear systematically due to the production layout.
It would be interesting to see how the different approaches (e.g. boosting, bagging) compare to each other across multiple datasets in a comprehensive comparison.
For instance, the proposed subsetting strategies (by grouped signals or common routes) are only basic ideas that could be analyzed in more detail and developed further.
In that sense, the propose method is still very elementary, and many variants could be further developed and explored.

%
%
\bibliographystyle{splncs04}
\bibliography{./references}

\end{document}